\documentclass[letterpaper]{article} 
\usepackage[draft]{aaai2026}  
\usepackage{times}  
\usepackage{helvet}  
\usepackage{courier}  
\usepackage[hyphens]{url}  
\usepackage{graphicx} 
\usepackage{microtype}
\usepackage{etoolbox}
\makeatletter\g@addto@macro\UrlBreaks{\do\-\do\.\do\/\do\_\do\?\do\&\do\=\do\%}\makeatother
\usepackage{natbib}  
\usepackage{caption} 
\usepackage{booktabs}
\usepackage{array}
\usepackage{tabularx}
\usepackage{amsmath}
\usepackage{amssymb}
\usepackage{multirow}
\usepackage{placeins}
\usepackage{xcolor}

\title{No Hindsight for LLM Fact-Checkers: Measuring Leakage Channels in Misinformation Detection}

\author{
    Kuan-Hua Wu Lu\textsuperscript{\rm 1},
    Yohanes Andre Setiawan\textsuperscript{\rm 1}
}
\affiliations{
    \textsuperscript{\rm 1}Georgia Institute of Technology\\
    Atlanta, Georgia, USA\\
    klu305@gatech.edu, ysetiawan3@gatech.edu
}

\begin{document}
\setcounter{secnumdepth}{2}
\raggedbottom
\fussy
\emergencystretch=1.5em
\makeatletter
\setlength{\@fptop}{0pt}
\setlength{\@fpsep}{8pt}
\setlength{\@fpbot}{0pt plus 1fil}
\renewcommand{\dbltopfraction}{0.9}
\setcounter{dbltopnumber}{2}
\makeatother
\AtBeginEnvironment{thebibliography}{%
  \small
  \raggedright\sloppy
  \setlength{\itemsep}{0pt}%
  \setlength{\parsep}{0pt}%
  \setlength{\bibsep}{0pt}%
}
\setlength{\parskip}{0pt}

\maketitle

\begin{abstract}
As automated fact-checking scales on social media, large language model (LLM) verdict scores can look stronger than warranted. One reason is that evaluations mix in information that was not knowable at claim time. Two channels are easy to conflate: outcomes memorized in pre-training and retrieved evidence published after the claim. Yet standard benchmarks rarely separate the two. In this study we measure both channels on AVeriTeC and QuanTemp++ by reconstructing point-in-time evidence conditions and probing for outcome information encoded in model representations. We find substantial evidence of parametric leakage, that can be hidden by the aggregate accuracy, while a simple representation bottleneck reduces this future leakage more efficiently than a mutual-information-based training penalty. We also find that allowing post-claim evidence inflates zero-shot accuracy by 6.3 points in AVeriTeC while the effect is negligible in QuanTemp++, where retrieval provides little post-claim evidence. These results show that misinformation benchmarks can overstate fact-checking performance when they do not account for what information was actually available at claim time.
\end{abstract}

\section{Introduction}
\label{sec:intro}

As automated fact-checking scales on social media to match the volume of
viral claims~\cite{Guo2022Survey}, recent systems put a large language
model (LLM) at the center of the pipeline, using it to read retrieved
evidence, write a rationale, or issue the verdict
itself~\cite{Chern2023Factool, Braun2025Defame}. Those systems are then
scored on benchmarks of claims that already have labels. It is a
convenient design, and it quietly hands the model something no deployed
fact-checker ever has: the future.

Consider the claim that COVID-19 was manufactured in a laboratory. In
April 2020, Facebook treated the claim as false and removed posts that
repeated it; on 27 May 2021 it reversed course and allowed the posts
back, because the lab-leak hypothesis had begun to attract serious
attention~\cite{FacebookLabLeak2021}. The claim never changed. Only what
was knowable about it did. A model scored on this claim today can reach
the right verdict two ways: by reasoning over what a fact-checker could
have read in April 2020, or by recalling how the story ended. A
benchmark that reports a single accuracy number cannot tell the
two apart, yet standard scores treat both paths as success. That
claim-time versus later-knowledge gap is what we measure.

Hindsight reaches a fact-checker through two channels that are easy to
conflate. The \emph{parametric} channel is outcome knowledge absorbed
during pre-training: the model has already read how the event resolved.
The \emph{temporal} channel is retrieved evidence published after the
claim, handed to the model by its own retrieval step. Existing practice
addresses neither directly. One line of work evaluates only on text that
postdates the training period~\cite{Lazaridou2021MindTheGap}, which is
clean but discards in-window capability on every widely used benchmark.
Another improves retrieval or evidence quality, including under
post-cutoff claims~\cite{Ou2026Aletheia}, but still does not date
retrieved pages against the claim.

In this study we measure both channels on AVeriTeC and
QuanTemp++~\cite{Schlichtkrull2023,QuanTempPP2025}, with a masking probe
on a frozen LLM, a trained width/penalty head, and a point-in-time
evidence split. Neither release records a clean publication date for every
evidence page: QuanTemp++ ships claim dates and noisy search-snippet
dates, and AVeriTeC ships archive snapshot times, so we recover dates
ourselves. We try the live source URL first and fall back to Wayback when needed,
parse on-page signals after stripping the archive prefix, and use the
archive timestamp only as a last-resort upper bound. That split yields a point-in-time view
with only pre-claim documents and the post-claim view a benchmark supplies
by default. On a frozen backbone we then compare a width bottleneck
against a mutual-information penalty on the realized
outcome~\cite{Cheng2020CLUB}.

Three findings follow. First, the parametric channel is real and hides
inside aggregate accuracy. Masking named entities (identity cues that can
anchor memorized outcomes) collapses supported-claim recall from 0.42 to
0.13 while overall accuracy moves only 0.03 points, and a pre-cutoff
control moves the opposite way. Second, narrowing representation width
reduces leakage more per point of macro-F1 than the mutual-information
penalty. Third, post-claim evidence raises zero-shot accuracy by $6.3$
points on AVeriTeC ($0.782$ point-in-time vs.\ $0.844$ post-claim), of
which $3.1$ is strict hindsight, while the gap is flat on QuanTemp++.
Point-in-time retrieval and width control reduce leakage but do not remove
it; fact-checkers should be judged on what they could have known at claim
time. We next turn to related work, then the measurement setup and results.
Both leakage channels, and both controls, are evaluated under that standard.

\section{Related Work}
\label{sec:related}

\paragraph{Evidence leakage in fact-checking.}
Several studies examine leakage inside a fact-checking pipeline, where
the evidence given to the model already contains an answer. Glockner
et al.~\shortcite{Glockner2022} show that in datasets such as
MultiFC~\cite{Augenstein2019}, the evidence often comes from the same
fact-checking article that states the ruling, so the model can copy it
instead of verifying the claim. Chrysidis et al.~\shortcite{Chrysidis2024}
filter leaked or low-quality evidence text, but their filter targets
content overlap rather than whether a page was published after the
claim. Particularly relevant to our evaluation are
AVeriTeC~\cite{Schlichtkrull2023} and
QuanTemp++~\cite{QuanTempPP2025}, which construct cleaner web-evidence
benchmarks to avoid that shortcut from the start; Thibault et
al.~\shortcite{Thibault2024} still find that leakage rates vary widely
across datasets. Neither release, however, records a publication date
for every page in its evidence pool, so the release alone does not
tell us which documents existed when the claim was made. El Lahib et
al.~\shortcite{ElLahib2026} show that live search-engine date filters still
return post-cutoff pages; we instead date the evidence already provided
with each benchmark claim. Prior work
describes, filters, or designs around leaked evidence; to the best of
our knowledge, none splits a standard benchmark into what was knowable
at claim time and what arrived later.

\paragraph{Temporal bias and lookahead in LLM evaluation.}
Language models can also exploit information unavailable at the moment
of prediction. Lazaridou et
al.~\shortcite{Lazaridou2021MindTheGap} find quality decays on text written
after the training cutoff, and Xu et
al.~\shortcite{Xu2024BenchmarkContaminationSurvey} trace how evaluation
data seeps into pre-training and inflates reported scores. He et
al.~\shortcite{He2025ChronologicallyConsistentLLMs} train only on text
available up to a given date; Karger et
al.~\shortcite{Karger2025ForecastBench} evaluate on genuinely future events
that no model could have memorized. 
Karnam et al. \shortcite{karnam2026} similarly highlight the importance of temporal realism in misinformation detection, using time-ordered evaluation to avoid assuming access to future social context that would be unavailable at deployment time. Particularly relevant to our probe
design is Zhu et al.~\shortcite{Zhu2022GeneralizingFuture}, who trace
detector errors to entity--label correlations that hold in the past
but break for future events. While they diagnose temporal shifts in
those correlations, we use entity masking to measure how much a
detector's representation still encodes the realized outcome, and we
contrast the same probe on a pre-cutoff control model that cannot have
memorized the later outcomes. To the best of our knowledge, prior work
detects or avoids the problem by changing the data or the model rather
than measuring residual outcome encoding in the representation.

\paragraph{LLM-based fact-checking.}
Hu et al.~\shortcite{Hu2024BadActor} find that LLMs make weak stand-alone
judges but useful advisors.
FacTool~\cite{Chern2023Factool} augments them with external tools, and
DEFAME~\cite{Braun2025Defame} couples an LLM with retrieved textual and
visual evidence. These pipelines treat recalled knowledge as extra
context and retrieve without a claim-date cutoff. They do not separate
outcomes memorized in pre-training from retrieved evidence published
after the claim.

\paragraph{Representation-level controls.}
A second family of tools tries to stop a representation from encoding
information it should not use. Width bottlenecks do this by capacity:
a narrower $R$ simply cannot carry as much. Information penalties do it
by objective, subtracting an estimate of dependence from the training
loss. Estimating mutual information in high dimensions is hard, and the
usual neural estimators are built to maximize dependence rather than
remove it. Nearest-neighbor methods~\cite{Kraskov2004EstimatingMI} and
lower bounds such as MINE~\cite{Belghazi2018MINE} cannot certify that a
dependence has shrunk; CLUB~\cite{Cheng2020CLUB} is a contrastive upper
bound, so it points in the direction a penalty needs. We compare both
knobs on the same frozen-backbone fact-checking head: a plain width
bottleneck and a conditional MI penalty on the realized outcome. Prior
work does not ask which of them actually removes outcome information
from that representation; when we compare them, width shrinks leakage
more than the penalty.

Closest to our setting are Glockner, Hou, and Gurevych~\shortcite{Glockner2022} and Zhu
et al.~\shortcite{Zhu2022GeneralizingFuture}: both diagnose a shortcut in
the data. To the best of our knowledge, no prior study jointly
measures parametric and temporal leakage on the same claims. In this
work we separate the channels with an entity-masking probe and
pre-cutoff control (parametric), a width-versus-MI comparison
(control), and a dated point-in-time versus post-claim split
(temporal) on AVeriTeC and QuanTemp++.


\section{Data}
\label{sec:data}

We measure both leakage channels on two public fact-checking benchmarks.
AVeriTeC~\cite{Schlichtkrull2023} provides real claims with human
labels, verdict text, claim dates, and web-evidence answers (NeurIPS
2023 Datasets and Benchmarks; CC BY-NC 4.0). Each answer stores a source URL
and a \texttt{web.archive.org} snapshot URL. QuanTemp++~\cite{QuanTempPP2025}
extends QuanTemp~\cite{Venktesh2024} with claim dates and evidence sources;
we apply the same cleaning, deduplication, and split logic on both; date
recovery differs by source (below). Neither release records a publication date
for every evidence page, so we recover dates ourselves and build two evidence
views per claim.

\subsection{Recovering evidence publication dates}
\label{sec:dates}

The point-in-time setting needs a publication date for every piece of
evidence; AVeriTeC provides only an archive timestamp inside each answer's
\texttt{web.archive.org} snapshot URL. That timestamp marks when the page was
archived, not when it was published, so we recover publication dates from the
page itself. We try the live source URL first and fall back to its Wayback capture
when needed, strip the archive prefix before parsing URL slugs, and read a
date using the signals in Table~\ref{tab:datetrust}, from most to least
reliable.\footnote{Some Wayback captures are served gzip-compressed; we detect
and decode them, since a garbled page is indistinguishable from one with no
date.} When a page has no on-page date, we keep the archive timestamp only as
an explicit upper bound.

\begin{table}[t]
\centering
\footnotesize
\caption{Publication-date signals on AVeriTeC binary claims, from most to least
reliable. ``Archive'' is an upper bound only, not a publication date.}
\label{tab:datetrust}
\begin{tabular}{@{}lp{0.42\columnwidth}rr@{}}
\toprule
Trust & Signal & Answers & Share \\
\midrule
slug & date in the URL path (\texttt{/2020/10/30/}) & 533 & 9.6\% \\
published & JSON-LD \texttt{datePublished}, & 1{,}603 & 28.8\% \\
 & \texttt{article:published{\allowbreak}\_time} metadata & & \\
meta & generic date metadata, \texttt{<time>} tags & 96 & 1.7\% \\
archive & snapshot timestamp (upper bound only) & 2{,}252 & 40.5\% \\
--- & no signal recovered & 1{,}081 & 19.4\% \\
\bottomrule
\end{tabular}
\end{table}

Across the 5{,}565 evidence answers on binary claims, 419 of the 2{,}232
with a trusted on-page signal (slug, published, or meta) postdate the claim
(18.8\%). We did not hand-audit a stratified sample of recovered dates
(Limitations). When a page carries several candidates, we keep the earliest
date in the most trusted tier (JSON-LD \texttt{datePublished} before generic
metadata).

\subsection{Point-in-time and leaked evidence views}
\label{sec:pit}

From the dated evidence we build the two views used in
Sections~\ref{sec:replevel} and~\ref{sec:temporal}. The
\emph{leaked} view is the cleaned evidence pool with no date filter. The
\emph{point-in-time} view $E(X)$ keeps an answer only if its recovered date
is non-empty and on or before the claim date. Archive bounds cut both ways:
a pre-claim snapshot proves the page existed then, but a post-claim snapshot
proves nothing about publication, so late bounds are dropped
(Table~\ref{tab:composition}). Of 1{,}337 binary claims with any
point-in-time evidence, 1{,}310 have a trusted on-page date and only 27 rely
on an archive bound alone; exclusion depends more on undated or late-bound
answers than on inclusion. The view is a lower bound on pre-claim evidence;
the temporal headline gap mixes that sparsity with strict hindsight
(Table~\ref{tab:decomp}). Claims with no qualifying answers keep an empty
point-in-time set rather than being dropped. Under either view we rank the
remaining answers by similarity to the claim and keep the top-$k$. Under
this rule, 1{,}078 of
2{,}385 binary train claims (45\%) and 259 of 410 dev claims (63\%) retain
at least one point-in-time answer; dev coverage is higher because later
claims carry more recoverable dates. On the chronological train+validation
split used in Section~\ref{sec:temporal}, 1{,}361 of 2{,}513 binary claims (54\%) have an empty
point-in-time set; Section~\ref{sec:temporal} reports that group separately.

\subsection{Processing and splits}
\label{sec:processing}

One shared pipeline produces the final records on both benchmarks. Labels map
to Supported, Refuted, and Mixed, with classes such as \emph{Not Enough
Evidence} discarded; the experiments here use only the binary subset (2{,}795
official claims on AVeriTeC). The pipeline removes evidence pointing back to
fact-checking sites, drops duplicates by URL and by near-identical text
(Jaccard $\ge 0.8$), drops claims with a bad date, a missing verdict, or
no evidence left, and collapses duplicate claim IDs to the earliest date.
AVeriTeC yields 3{,}010 processed records across all three
label classes on the official split; chronological splitting removes 11
records whose verdict text duplicates train, leaving 2{,}999 claims in that
pool (2{,}513 binary train+validation for temporal analysis). Next to the official splits we
built a chronological 80/10/10 split by claim date so training always
precedes validation and test, checking that no record appears in both train
and test and removing validation or test records whose verdict text also
appears in train. Data are public archived web pages with
no private user information (AVeriTeC CC BY-NC 4.0; QuanTemp++ under its
release terms). At camera-ready we will release the date cache (URL, recovered
date, trust tier, and signal), split indices, generated $K$ text, and code;
we do not redistribute recovered HTML, which includes third-party, social-media,
and paywalled pages outside those licenses.

\subsection{QuanTemp++}
\label{sec:secondarydataset}

QuanTemp++ shares our label mapping, fact-check-site filter, deduplication,
and chronological split, but dates come mainly from URL slugs and search
snippets in the release; undated URLs use the same page-date recovery as
above (archive bounds almost never appear). Repeated scraper placeholder
dates in search snippets are dropped. Its point-in-time view therefore uses
a different dating instrument than AVeriTeC's, not a matched corpus. The
chronological split has 4{,}827 train and 710 validation binary claims; only
4.1\% of all evidence answers postdate the claim (5.8\% of the leaked top-5
in Table~\ref{tab:composition}), yet 70.4\% of claims still retrieve the same
top-5 under both views, so the flat temporal gap
reflects low post-claim volume and limited rank change, not a controlled
match. The validation split is 5.5\% Supported, so Supported-side estimates
are noisy.

\subsection{Outcome statements $K$}
\label{sec:koutcome}

The leakage penalty needs $K$, an embedding of the claim's realized outcome.
On AVeriTeC we prompt a local LLM (Qwen3-4B) with the claim as a focus hint
and the fetched fact-check article, but never the verdict label, asking for a
short neutral statement of how events unfolded and banning verdict and
negation words. Unreadable pages yield no article-based $K$; we later tried to
regenerate 43 such cases from the dataset \texttt{justification} field
(37 recovered; 6 remain without $K$), which we otherwise never feed as claim
or evidence input. QuanTemp++ has no
article URLs, so its $K$ statements are generated from the fact-checker
\texttt{justification} text under the same prompt---a limitation for the
representation-level analysis on that corpus (Limitations). Violations are flagged rather than filtered,
because negation is often inherent to the outcome. In our generation run on
the 3{,}010 processed AVeriTeC records across all three label classes, 66
had unreadable articles; of the 2{,}944 statements we could generate,
1{,}216 (41\%) carry a flag, concentrated on Refuted claims (52\% vs.\ 17\%
on Supported), so part of what the representation-level probe measures is label information
rather than outcome knowledge. Generated $K$ text will be released at
camera-ready.

\section{Methodology}
\label{sec:method}
\begin{figure*}[!t]
\centering
\includegraphics[width=0.95\textwidth]{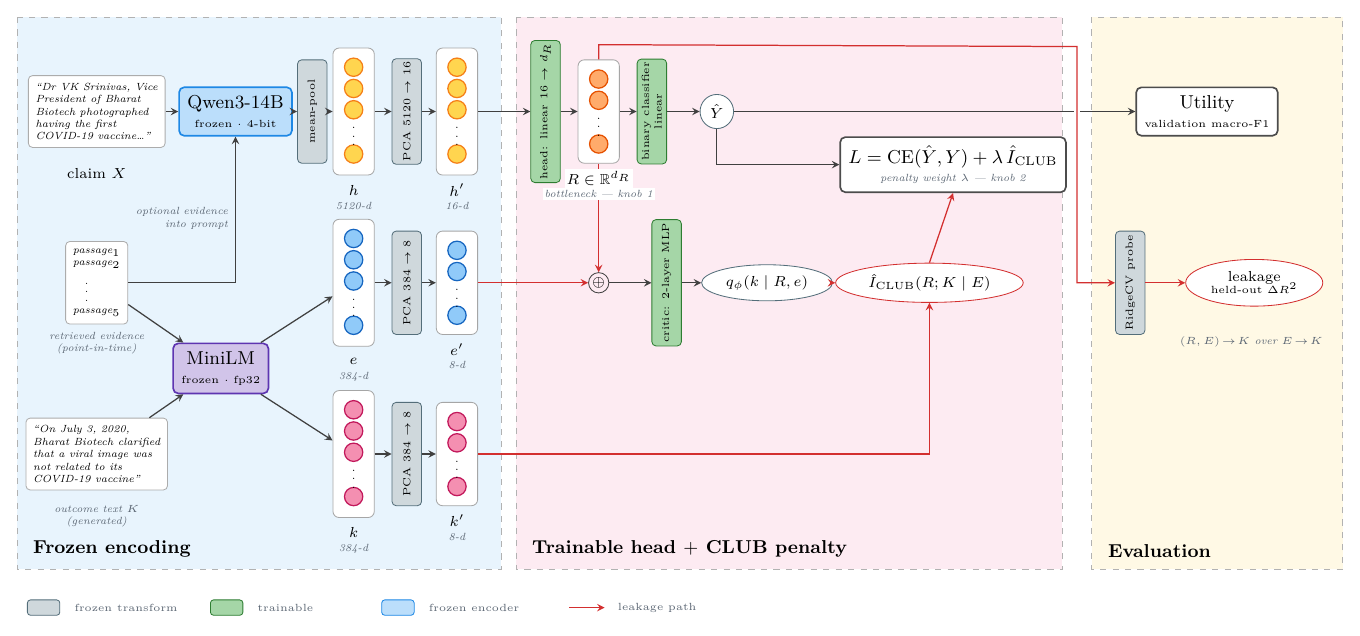}
\caption{Trained representation pipeline. Two knobs---bottleneck width $d_R$ and penalty weight $\lambda$---are studied against future-information leakage.}
\label{fig:architecture}
\end{figure*}
\subsection{Problem setup and notation}
For a claim $X$ we retrieve point-in-time evidence $E(X)$ and form a representation
$R=f\!\big(X,E(X)\big)\in\mathbb{R}^{d}$, from which a classifier predicts the binary
label $Y$ (true/false). We also define $K$, an embedding of the realised
outcome (Section~\ref{sec:koutcome}). Since $K$ is derived from post-claim
adjudication material excluded from retrieval, it carries future outcome
information we do not want $R$ to encode beyond what the selected evidence
already supports. Figure~\ref{fig:architecture} shows the full pipeline.

\subsection{Evidence retrieval}
The evidence $E(X)$ in $R=f(X,E(X))$ comes from the evidence each benchmark
already provides for each claim, given as question--answer pairs with source
links. We do not retrieve from the open web. We score and select within this
provided set. We first apply the point-in-time rule of Section~\ref{sec:pit}.
We then embed the claim and each remaining answer with a pretrained sentence
encoder (\texttt{all-mpnet-base-v2})~\cite{Reimers2019SBERT}, rank the answers
by their similarity to the claim, and keep the top-$k$ (or all of them when a
claim has fewer than $k$) as the point-in-time evidence set $E(X)$. This
encoder is used only for ranking; a separate encoder
(\texttt{all-MiniLM-L6-v2})~\cite{Reimers2019SBERT} embeds the selected
evidence text as $E$, which serves as the conditioning variable in
$I(R;K\mid E)$. Conditioning on $E$ keeps the penalty aimed only at the
outcome information beyond what the selected evidence already supports.

\subsection{Learning objective}
We train $R$ to keep the veracity signal while avoiding using future-outcome information by solving the function
($I(\cdot;\cdot)$ denotes mutual information):
\begin{equation}
\max_{f}\;\;
\mathcal{L} \;=\; \underbrace{I(R;Y\mid X)}_{\text{utility}} \;-\;
\lambda\,\underbrace{I(R;K\mid E)}_{\text{leakage}},
\qquad \lambda>0 .
\end{equation}
The utility term is maximized by minimizing the cross entropy loss in the training
classifier. The leakage term is minimized through the
conditional-CLUB upper bound~\cite{Cheng2020CLUB},
\begin{equation}
\begin{split}
I(R;K\mid E) \;\le\;{}& \mathbb{E}_{p(r,k,e)}\!\big[\log q_\phi(k\mid r,e)\big] \\
&{}-\; \mathbb{E}_{p(e)}\,\mathbb{E}_{p(r\mid e)\,p(k\mid e)}\!\big[\log q_\phi(k\mid r,e)\big].
\end{split}
\end{equation}
where $q_\phi(k\mid r,e)$ approximates $p(k\mid r,e)$.

\subsection{Representation of feature $R$}
We use a moderately sized \textbf{Qwen3-14B} model, instruct the model to run
in non-thinking mode and keep weight frozen. For each claim we feed the claim text together 
with its evidence and mean-pool the final-layer hidden states into a fixed feature vector. 
Because the backbone weights never change, these features are identical on every pass, so we 
compute them once and cache them. A small trainable "representation" head then translates the 
cached features to $R\in\mathbb{R}^{d}$.

\subsection{Outcome $K$ and evidence $E$ encoders}
$K$ and $E$ are encoded by a separate, fixed sentence encoder (\texttt{all-MiniLM-L6-v2}), not the Qwen3-14B backbone model. Using a different encoder helps the penalty measures real
outcome leakage rather than encoder similarity. We reduce both $K$ and $E$ to lower ($8$) dimensions with PCA (a separate projection for each, fit on the training split and applied to all splits), to help training stability given the moderate data size we have

\subsection{Training}
We optimize the model through alternating optimization of two objectives:
\begin{enumerate}
\item \textbf{Estimate the leakage.} With the head and classifier frozen, update
$q_\phi(k\mid r,e)$ to \emph{maximize} $\sum_i \log q_\phi(k_i\mid r_i,e_i)$ --- make the helper
predict each outcome $k_i$ from $(r_i,e_i)$ as well as possible, so the CLUB value is an
accurate estimate of $I(R;K\mid E)$.
\item \textbf{Learn the representation $R$.} With $q_\phi$ frozen, update the head and
classifier to minimize
$\mathrm{CE}\big(Y,\,q_\psi(R)\big) + \lambda\,\widehat{I}_{\text{CLUB}}(R;K\mid E)$. The
cross-entropy term drives $R$ to predict the label, while the CLUB term
penalizes $R$ for encoding realized-outcome information.
\end{enumerate}

\section{Analysis and Results}
\label{sec:analysis}
\subsection{Parametric leakage}
\label{sec:parametric}
We first test whether a modern LLM's judgment is based on memorized
outcomes from pre-training or on its ability to verify a claim from its
content. We evaluate this with a two-pass test on the 2{,}795 binary
AVeriTeC claims without retrieval. In the first pass, the model classifies the
original claim and in the second pass, we mask the named entities, so that the wording and structure remain intact, but the name entities
a model could use to recall a specific outcome are removed.

The masking uses spaCy NER. Claims where NER finds no entity (241
of 2{,}795) are dropped from this probe, leaving 2{,}554 claims (819
supported). The model is prompted to answer True or False from its own
knowledge, and $P(\text{true})$ is the next-token probability mass the model predicts ``True'' against ``False''.  We expect the model judgment to be roughly unchanged if the model were verifying the claim from its content alone. A collapse concentrated on the claims whose outcomes are memorable is likely hint of parametric leakage.

\subsubsection{Results}

\begin{table}[t]
\centering
\small
\caption{Entity-masking probe on Qwen3-14B, 2{,}554 entity-bearing binary
AVeriTeC claims, no evidence shown. Majority is always-Refuted
(1{,}735/2{,}554).}
\label{tab:rq1}
\begin{tabular}{@{}lrrrr@{}}
\toprule
Metric & Original & Masked & $\Delta$ & Majority \\
\midrule
Accuracy & 0.710 & 0.678 & $-0.032$ & 0.679 \\
Macro-F1 & 0.640 & 0.502 & $-0.138$ & 0.405 \\
Supported recall & 0.421 & 0.131 & $-0.291$ & 0.000 \\
Refuted recall & 0.846 & 0.937 & $+0.091$ & 1.000 \\
\bottomrule
\end{tabular}
\end{table}

Table~\ref{tab:rq1} shows the headline pattern. The Supported-recall drop is
0.291 (0.421 $\rightarrow$ 0.131). Two things stand out in Table~\ref{tab:rq1}. First, raw accuracy hides the
leakage: the gap is only 0.032, because the two recall movements nearly
cancel on a set that is 68\% Refuted. Masked accuracy (0.678) matches
the always-Refuted majority (0.679). An accuracy report alone
would conclude the model is robust to masking. However, the per-class recalls show
the opposite. Second, the movement is asymmetric in the direction
memorization predicts. The model recognizes true 2020 outcomes only while
their entity anchors are present. When the anchors are gone it does not
reason, it falls back to its ``False'' prior.

Table~\ref{tab:examples} provides examples. Claims whose outcomes were
widely reported in 2020 go from full confidence to rejection once the
entities are hidden.

\begin{table}[t]
\centering
\small
\caption{Supported claims where masking flips full confidence into
rejection ($P(\text{true})$, original $\rightarrow$ masked).}
\label{tab:examples}
\begin{tabular}{@{}p{0.58\columnwidth}r@{}}
\toprule
Claim (original / masked) & $P(\text{true})$ \\
\midrule
\emph{Russia has developed a new vaccine to fight COVID-19.} &
$1.00 \rightarrow 0.00$ \\
\emph{[GPE\_1] has developed a new vaccine to fight [PERSON\_1].} & \\
\addlinespace
\emph{American rapper Kanye West is running for President of the United
States in the 2020 elections.} & $1.00 \rightarrow 0.00$ \\
\emph{[NORP\_1] rapper [PERSON\_1] is running for President of [GPE\_1] in
the [DATE\_1] elections.} & \\
\bottomrule
\end{tabular}
\end{table}

\subsubsection{Pre-cutoff control model}
\label{sec:control}

Masking removes information, not only memorization anchors. A masked claim
is naturally harder to verify, so part of the drop could be information loss
rather than leakage. To isolate the memorization component we run the
identical probe (same claims, same masking, same instruction text as a plain
prompt, same scoring) on a control model whose pre-training predates the
claim outcomes: GPT-2-XL, trained on WebText, which was
collected through December 2017 and so predates 97.5\% of the claims.
A model that never saw an outcome has nothing to recall through the entity anchors, so its
masking drop estimates the pure information-loss component, and the excess drop of the in-window
model over the control is the memorization component. A control that shows
no drop only because of underperformance would prove nothing, so
Table~\ref{tab:control} also reports the control's unmasked performance.

\begin{table*}[!t]
\centering
\footnotesize
\caption{The identical probe on a pre-cutoff control and a second in-window
model. Recall figures and CIs
on the entity-bearing subset; $\Delta P(\text{true})$ is the confidence
change on gold-Supported claims (positive = masking lowers confidence).
Both in-window models collapse toward ``False'' but the control moves in the
opposite direction.}
\label{tab:control}
\begin{tabular}{@{}llcccc@{}}
\toprule
Model & Training data & Acc. & Supp.\ recall
(o$\rightarrow$m) & Recall drop
[95\% CI] &
$\Delta P(\mathrm{true})$
[95\% CI] \\
\midrule
Qwen3-14B & in-window (2020) & 0.710 & $0.421\!\rightarrow\!0.131$ &
$+0.291$ $[.258,.322]$ & $+0.279$ $[.250,.309]$ \\
phi-4 & in-window (2020) & 0.717 & $0.277\!\rightarrow\!0.039$ &
$+0.238$ $[.210,.268]$ & $+0.250$ $[.227,.274]$ \\
GPT-2-XL & WebText $\leq$2017 & 0.606 & $0.394\!\rightarrow\!0.686$ &
$-0.292$ $[-.322,-.260]$ & $-0.087$ $[-.092,-.082]$ \\
\bottomrule
\end{tabular}
\end{table*}

The control does not show the collapse. GPT-2-XL accuracy is weak
(unmasked 0.606, below the 0.679 majority in Table~\ref{tab:rq1}) but
still splits both classes away from the extremes, and masking changes its verdicts substantially,
interestingly in the opposite direction of Qwen3-14B. Its Supported recall rises and its confidence on gold-True claims
goes up. For a model with nothing to recall, replacing entities with
placeholders produces a drift toward ``True'' that has nothing to do with
the claim content. The
difference-in-differences on the paired entity-bearing claims is $+0.582$
(95\% CI $[+0.533, +0.630]$). Since the GPT control's
drop is below zero, the information loss alone does not produce a collapse
toward ``False''. The in-window drop of 0.291 is mostly attributable
to recall of memorized outcomes rather than to the claims becoming harder to
verify, and Table~\ref{tab:ddp} separates the two parts.

\subsubsection{Robustness checks}
\label{sec:rq1robust}

\begin{table}[t]
\centering
\footnotesize
\caption{The confidence change under masking, split by gold label,
with $\Delta\Delta P$ the gold-True minus the gold-False change (positive =
masking lowers confidence). A label-blind drift moves both classes equally
and cancels in $\Delta\Delta P$.}
\label{tab:ddp}
\begin{tabular}{@{}lccc@{}}
\toprule
Model & gold-True & gold-False & $\Delta\Delta P$ [95\% CI] \\
\midrule
Qwen3-14B & $+0.279$ & $+0.089$ & $+0.190$ $[.156,.225]$ \\
phi-4 & $+0.250$ & $+0.075$ & $+0.175$ $[.150,.202]$ \\
GPT-2-XL & $-0.087$ & $-0.096$ & $+0.009$ $[.002,.015]$ \\
\bottomrule
\end{tabular}
\end{table}

The collapse is not specific to one model family. The identical probe
on phi-4 (14B), a second in-window model with a different training pipeline,
shows the same pattern (Table~\ref{tab:control}). Its supported recall falls
from 0.277 to 0.039 while its refuted recall rises, and raw accuracy again
hides the movement.

Masking could also simply push each model toward its own prior, Qwen
toward ``False'' and GPT-2 toward ``True'', and the gap in
Table~\ref{tab:control} would widen without any memorization. We check the distribution shift across labels to confirm future leakage. For models simply falling back to their prior, we expect the shift
between the two labels cancels out. However, memorized outcomes do
not cancel. Table~\ref{tab:ddp} reports the difference. For both in-window
models the confidence loss is about three times larger on gold-True claims
than on gold-False claims. For the pre-cutoff control the shift is
label-blind to within 0.01. The label conditioned signature of the collapse is the
memorization behavior.

\subsection{Representation-level controls}
\label{sec:replevel}
The training objective (Section~\ref{sec:method}) adds a penalty
$\lambda \, I(R; K \mid E)$ on the representation $R$, the vector constructed using claim and evidence (details below). The penalty's job is to keep $R$
from relying on the realized outcome $K$ beyond what the evidence
already supports. We next ask whether this design is effective: as $\lambda$ grows, does
the leakage actually fall, and does it come at a trade-off against accuracy?

The measurement of leakage needs to be outside of the training process to ensure the penalty term is not grading its own work. In our experiment, we use a helper model
to predict $K$ from $R$ and $E$, another predicts $K$ from $E$ alone, both after training, and the
gain in held-out $R^2$ is the leakage number. It is trying to measure how much $R$ reveals about the
outcome beyond what the evidence supports. Both helpers are linear ridge regressions, fit on the training split and scored on the validation split. We pick the regularization strength by cross-validation.

\subsubsection{Setup}

We run the experiment on both datasets of Section~\ref{sec:data}. For AVeriTeC we start
from the same 2{,}795 binary claims as the parametric probe. The chronological split removes
near-duplicates while holding out a test split, leaving 2{,}248 train and 265 validation claims (2{,}513 binary train+validation). The penalty additionally requires a generated outcome statement $K$, so the leakage grid uses the 2{,}209 train and 261 validation claims that have $K$ (2{,}470 of those 2{,}513). Accuracy at $\lambda=0$ in Section~\ref{sec:temporal} is on all 265 validation claims, which do not need $K$ at inference.
QuanTemp++ gives 4{,}827 train and 710 validation claims. We remind the readers that the validation split for QuanTemp++ is 5.5\% Supported (39 claims). Therefore the supported-side estimate can be noisy.

The representation $R$ is built from the claim in four steps. First, a claim
$X$, alone or with its top-5 point-in-time evidence, is formatted as a single
prompt and passed through the frozen Qwen3-14B backbone. Second, the
final-layer hidden states are mean-pooled into a 5{,}120-dimensional vector.
Third, a PCA fit on the training split reduces it to 16 dimensions, keeping
59--61\% of the feature variance on both datasets. Fourth, a small trainable
head, a linear map or a shallow MLP chosen per dataset
(Appendix~\ref{app:configs}), projects those 16 dimensions to
$R \in \mathbb{R}^{d_R}$. A linear classifier predicts the label from $R$.
Only the fourth step is trainable.

The outcome statement $K$ and the evidence text $E$ are embedded by a
separate frozen sentence encoder (all-MiniLM-L6-v2) and reduced to 8 PCA
dimensions each, so the leakage term measures outcome dependence rather than
similarity between two views of the same encoder. These embeddings keep
52\% and 24\% of their variance on AVeriTeC and 27\% and 22\% on QuanTemp++,
to avoid overfitting given our data size.

The head is selected for the smallest train--validation gap that holds
validation F1, using the config most prone to overfitting ($\lambda{=}0$,
$d_R{=}32$, Appendix~\ref{app:configs}), so the model is scored on a
representation that generalizes. Training alternates between two updates. First the CLUB critic
$q_\phi(k \mid R, e)$ is fit to predict the outcome embedding from the
current representation and evidence. Its fit yields the estimate
$\hat{I}_{\mathrm{CLUB}}$ of $I(R; K \mid E)$. Then, with the critic held
fixed, the head and classifier take a step on the combined loss: the
cross-entropy classification loss plus
$\lambda\,\hat{I}_{\mathrm{CLUB}}$. Alternating keeps the leakage estimate
current as the representation changes. The width $d_R$ is the objective's
second knob: it is the number of
dimensions the classifier and the critic see, and at $\lambda{=}0$ it is a
plain bottleneck with no penalty. Every cell is trained with five seeds
varying only the initialization, sweeping
$\lambda \in \{0, 0.01, 0.02, 0.04, 0.06, 0.1, 0.5\}$ against
$d_R \in \{8, 2, 1\}$. Cells where the penalty appears to help are re-run
with 20 paired seeds (Appendix~\ref{app:details}).

\subsubsection{Results}

\begin{figure*}[!t]
\includegraphics[width=0.95\textwidth]{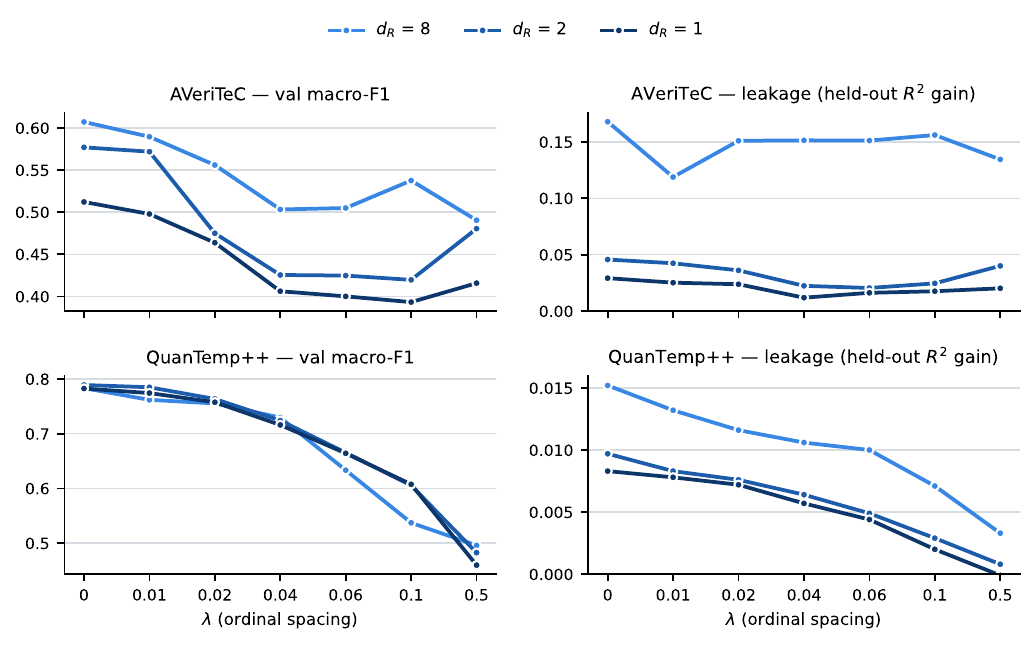}
\caption{Validation macro-F1 (left) and leakage (right) as the penalty
$\lambda$ grows, on AVeriTeC (top) and QuanTemp++ (bottom) with top-5
point-in-time evidence, one line per width $d_R$. F1 falls as $\lambda$ grows
for at most a marginal leakage return. The width, not the penalty, is what
moves leakage. Means over 5 seeds.}
\label{fig:rq2}
\end{figure*}

\begin{table}[t]
\centering
\footnotesize
\caption{AVeriTeC: the $\lambda{=}0$ grid with and without retrieved
point-in-time evidence. Leakage is the held-out $R^2$ gain of predicting $K$
from $(R, E)$ over $E$ alone. Means over 5 seeds. In the claim-only setting the
conditioner $E$ is a constant, so its leakage column reads unconditional
outcome decodability.}
\label{tab:rq2}
\begin{tabular}{@{}rrrrr@{}}
\toprule
& \multicolumn{2}{c}{top-5 point-in-time} & \multicolumn{2}{c}{claim-only} \\
\cmidrule(lr){2-3} \cmidrule(l){4-5}
$d_R$ & F1 & leakage & F1 & leakage \\
\midrule
 8 & 0.607 & 0.168 & 0.626 & 0.268 \\
 2 & 0.577 & 0.046 & 0.582 & 0.086 \\
 1 & 0.512 & 0.029 & 0.492 & 0.057 \\
\bottomrule
\end{tabular}
\end{table}

\begin{table}[t]
\centering
\footnotesize
\caption{QuanTemp++: the $\lambda{=}0$ grid with and without retrieved
point-in-time evidence, as Table~\ref{tab:rq2}. The claim-only leakage is
further inflated by outcome statements that restate the claim.}
\label{tab:qt}
\begin{tabular}{@{}rrrrr@{}}
\toprule
& \multicolumn{2}{c}{top-5 point-in-time} & \multicolumn{2}{c}{claim-only} \\
\cmidrule(lr){2-3} \cmidrule(l){4-5}
$d_R$ & F1 & leakage & F1 & leakage \\
\midrule
 8 & 0.783 & 0.015 & 0.730 & 0.392 \\
 2 & 0.789 & 0.010 & 0.731 & 0.307 \\
 1 & 0.783 & 0.008 & 0.730 & 0.288 \\
\bottomrule
\end{tabular}
\end{table}

We find increasing $\lambda$ does not effectively shrink leakage in both
datasets. Figure~\ref{fig:rq2} shows the sweep: at every width, F1 falls as
$\lambda$ grows for at most a marginal leakage return
(Appendix~\ref{app:details}).

What does shrink the leakage is the representation width $d_R$. Narrowing
the head removes far more leakage per unit of F1 than any $\lambda$
(Tables~\ref{tab:rq2} and~\ref{tab:qt}), and no penalized cell beats the
$\lambda{=}0$ bottleneck at equal utility.

We read this as a statement about where the leakage lives. Most of it is
carried in directions of $R$ that the classifier does not use, so shrinking
the width discards it at almost no cost. A penalty that must fight the
utility term for the same directions cannot do better. The remainder, about
$0.03$ of $R^2$ on AVeriTeC and less on QuanTemp++, survives every width and
every penalty. It lies inside the directions the prediction itself uses,
where a claim's verdict and its realized outcome overlap, so no
representation-level tool removes it without removing the prediction.

Both datasets are evaluated at the same two evidence settings, claim-only
and top-5 point-in-time. Claim-only is the higher-utility setting on AVeriTeC at $d_R \in \{8, 2\}$,
while evidence adds about 0.06 F1 on QuanTemp++
(Tables~\ref{tab:rq2} and~\ref{tab:qt}). The penalty helps only in
QuanTemp++'s claim-only setting within the same width, where there is no evidence to condition on
(Appendix~\ref{app:details}). In the settings with evidence, the bottleneck
remains the better tool.

\subsection{Temporal evidence leakage}
\label{sec:temporal}
The model keeps outcome knowledge in its weights (Section~\ref{sec:parametric}). Here we study the temporal channel on our trained detector alongside the parametric one. Section~\ref{sec:replevel} found that only the width bottleneck $d_R$ lowers the leakage probe, while the penalty $\lambda$ mostly costs accuracy; both knobs still change how much the head relies on evidence rather than on memory, which we reuse here. The temporal channel is evidence dated after the claim that retrieval brings in, and only point-in-time retrieval can remove it. We ask how this channel affects the trained detector, and how pushing off the parametric shortcut changes that sensitivity.

\begin{figure*}[t]
\centering
\includegraphics[width=0.82\textwidth]{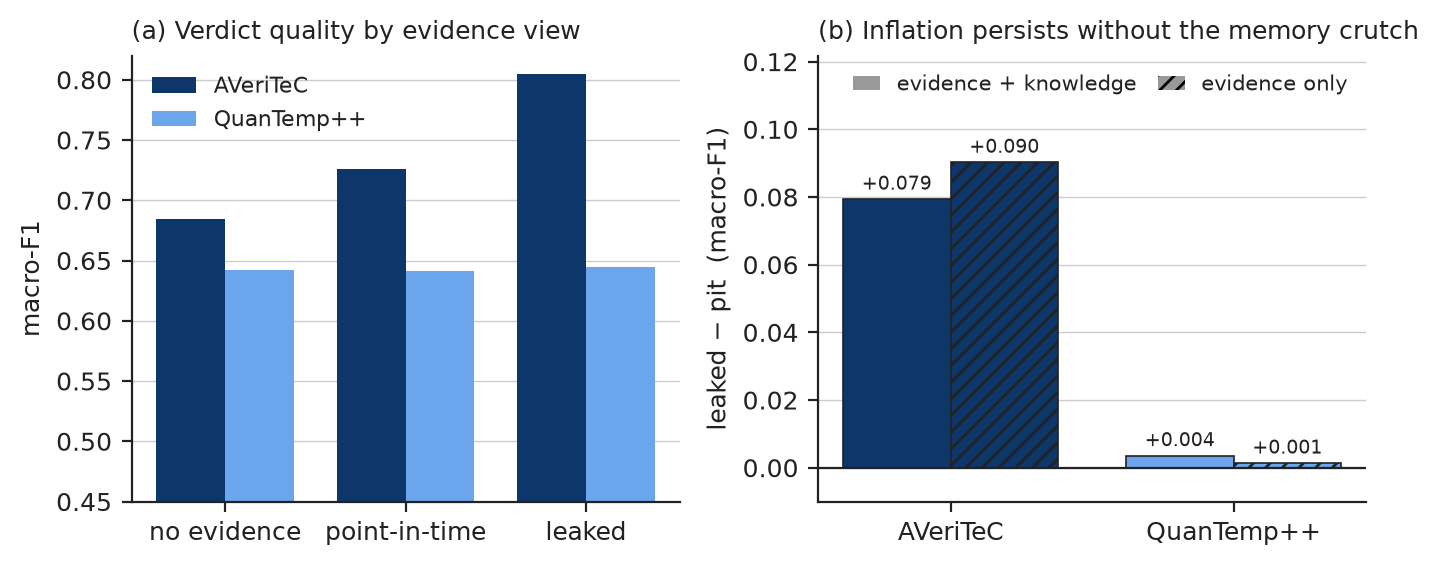}
\caption{Zero-shot Qwen3-14B verdicts. (a) macro-F1 by evidence view: on AVeriTeC quality
rises from no-evidence to point-in-time to leaked; on QuanTemp++ it is flat, because its retrieval surfaces almost no post-claim evidence (Table~\ref{tab:composition}). (b) leaked $-$ point-in-time inflation under the default prompt and an evidence-only prompt; the gap grows when the memory fallback is removed. Bars are on the full binary claim sets
($n=2{,}513$ / $5{,}537$).}
\label{fig:zeroshot}
\end{figure*}

\begin{figure*}[t]
\centering
\includegraphics[width=0.82\textwidth]{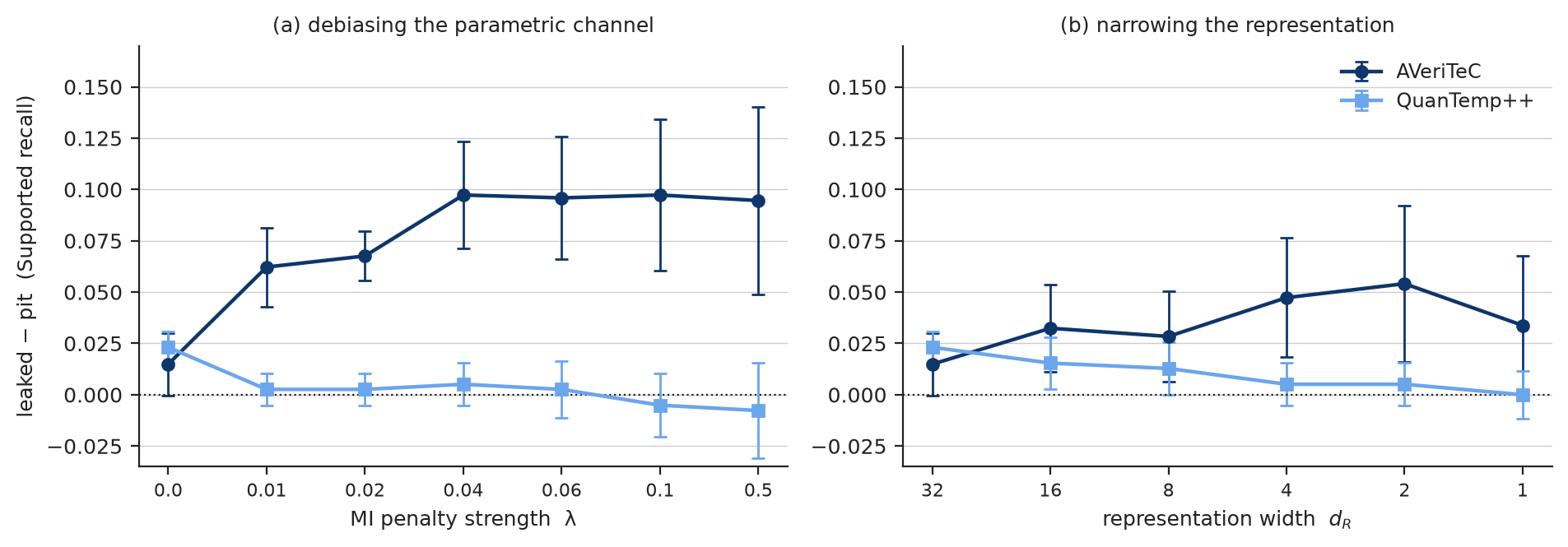}
\caption{Leaked $-$ point-in-time Supported-recall of the trained head vs.\ (a) $\lambda$ at $d_R{=}32$ and (b) width $d_R$ at $\lambda{=}0$ (Section~\ref{sec:replevel}). On AVeriTeC the gap is near zero while the parametric shortcut remains, then re-emerges; on QuanTemp++ it falls toward zero (Table~\ref{tab:composition}). Means over 10 seeds, $\pm 1$ SD.}
\label{fig:knobs}
\end{figure*}

\noindent\textbf{The temporal leakage is present in the retrieved data.} We first check that the leakage is really there, using the same backbone and top-$k$ retrieval as our detector. We give Qwen3-14B a claim and its top-5 evidence and ask it for a Supported or Refuted answer, with no training. Because this probe is zero-shot, we can test it on every binary claim in the chronological train and validation splits ($n=2{,}513$ of the 2{,}795 binary claims on AVeriTeC; the test split stays held out) instead of only the validation split, which gives it more statistical power. In the point-in-time view, claims with no dated pre-claim evidence keep an empty evidence set, so no claim is dropped. When we move from point-in-time to leaked evidence, accuracy goes up by $6.3$ points (95\% CI $[4.9, 7.8]$), macro-F1 by $7.9$ ($[6.1, 9.9]$), and Supported-recall by $10.4$ ($[7.0, 14.1]$), and all three gaps are significant (accuracy by McNemar $p<0.001$, and every bootstrap interval excludes zero, Figure~\ref{fig:zeroshot}). Two checks show that the gain comes from the leaked evidence and not from the model's own prior knowledge. First, when we tell the model to use only the given evidence, the gap gets even bigger (macro-F1 inflation $7.9$ to $9.0$). Second, the same model with no evidence at all reaches only $75.3\%$ accuracy, well below both evidence conditions, so prior knowledge alone does not account for its verdicts. The temporal leakage is therefore a property of the retrieved data, and any model that reads the evidence can use it.

\noindent\textbf{Part of this gap is missing evidence, not hindsight.} The $6.3$ point gap covers two different situations. For $54\%$ of the claims our date recovery finds no pre-claim evidence at all (Section~\ref{sec:pit}), so the point-in-time view gives the model an empty evidence set, and the comparison measures what a fact-checker loses when nothing was available yet. For the other $46\%$ both views have evidence and only the publication dates differ, which is hindsight in the strict sense. Table~\ref{tab:decomp} splits the two. The gap is $9.0$ accuracy points for the first group and $3.1$ points for the second, and both are significant (McNemar $p<10^{-4}$). We therefore read the $6.3$ points as how much a leaked benchmark overstates performance overall, and the $3.1$ points as an upper bound on the part that comes from hindsight, since the point-in-time view also drops evidence we simply could not date. On QuanTemp++ nearly every claim has pre-claim evidence, so its flat result is not caused by availability.

\begin{table}[t]
\centering
\footnotesize
\caption{The zero-shot accuracy gap on AVeriTeC, split by whether any pre-claim evidence
exists for the claim. Where none exists, the point-in-time view gives the model an empty
evidence set. Both gaps are significant.}
\label{tab:decomp}
\begin{tabular}{@{}lcccc@{}}
\toprule
Claims & $n$ & PIT & Leaked & $\Delta$ \\
\midrule
No pre-claim evidence  & 1{,}361 & 0.746 & 0.835 & $+0.090$ \\
Has pre-claim evidence & 1{,}152 & 0.824 & 0.855 & $+0.031$ \\
\addlinespace
All claims             & 2{,}513 & 0.782 & 0.844 & $+0.063$ \\
\bottomrule
\end{tabular}
\end{table}

\noindent\textbf{The detector hides this leakage until the parametric shortcut is removed.} At $\lambda=0$ on chronological validation ($n=265$ AVeriTeC, the full validation set; $n=710$ QuanTemp++) the trained pipeline shows almost none of the prompted model's inflation: a head trained on leaked features and one trained on point-in-time features differ by $-1.0$ accuracy points on AVeriTeC and near zero on QuanTemp++, with Supported-recall gaps of $+3.5$ and $+2.1$. That is a property of the pipeline, not an absence of leakage. The head leans on the parametric channel and barely reacts to which evidence view it is given, which is also why claim-only beats the evidence setting for it on AVeriTeC (Table~\ref{tab:rq2}) while the prompted model gains from that same evidence (Figure~\ref{fig:zeroshot}). Pushed off the shortcut it has to read the evidence, and the temporal leakage already sitting there starts to show (Figure~\ref{fig:knobs}): on AVeriTeC the leaked minus point-in-time Supported-recall gap rises from $1.5\%$ to $9.7\%$ as $\lambda$ grows and to $5.4\%$ as $d_R$ narrows to $2$, while on QuanTemp++ both knobs push it from $2.3\%$ toward zero. (The $1.5\%$ start is below the $+3.5$ above because that comparison trains a separate head per evidence view, while this sweep swaps the view only at test time.) The peaks sit inside the seed spread, so we read the direction and not a best value; that two unrelated knobs move the gap the same way makes it a property of the evidence rather than of one intervention. This squares with Section~\ref{sec:replevel}: the penalty barely changes what $R$ encodes about the outcome, but it does change what the head reads, so its effect surfaces here as evidence sensitivity rather than as a lower leakage number.

\noindent\textbf{The effect only appears in datasets where hindsight is actually retrieved.} On QuanTemp++ the zero-shot inflation is not significant for accuracy or macro-F1 (accuracy gap $0.05\%$, McNemar $p=0.87$). Table~\ref{tab:composition} explains why. On AVeriTeC the point-in-time filter removes $68\%$ of the leaked top-5 evidence, and only $19\%$ of claims get the same evidence in both views, so there is real room for hindsight to act. On QuanTemp++ $91\%$ of the leaked evidence would also pass the point-in-time filter, and $70\%$ of claims get the same top-5 either way, so the two settings are almost the same experiment and very little can leak. The size of the inflation follows how much post-claim evidence the retrieval brings in, which depends on how the dataset was built and not on the model.

Taken together, these experiments show that temporal evidence leakage is present in the retrieved data and that a model can use it if it reads the evidence, that our trained detector is safe from it at $\lambda=0$ only because it does not use the evidence much, and that reducing the parametric channel brings the temporal leakage back. \emph{Representation debiasing and point-in-time retrieval are complementary: pushing the head off its parametric shortcut exposes the temporal leakage that only point-in-time retrieval removes.}

\section{Limitations}
\label{sec:limitations}
First, we cannot show that the outcome information is fully removed. Most of our claims predate the backbone's training cutoff, and because the backbone stays frozen and only a small head is trained on its features, those frozen weights set a floor on what any representation-level tool can remove. Both the training penalty and the held-out probe also rely on estimating mutual information in high dimensions, and Section~\ref{sec:replevel} finds a residual of about $0.03$ held-out $R^2$ that survives every width and every penalty we tried, sitting in the directions the classifier itself uses.

Second, our point-in-time setting depends on recovering a publication date for each piece of evidence, and this coverage is incomplete. We did not hand-audit a stratified sample of recovered dates. Many answers stay undated---social-media and paywalled pages are common cases---and late archive bounds are dropped conservatively, so all are excluded from the point-in-time view, not only pages we can prove were published after the claim. This is why $54\%$ of the binary AVeriTeC claims end up with no pre-claim evidence at all, so part of what we measure as inflation is missing evidence rather than hindsight. Section~\ref{sec:temporal} reports the two cases separately for this reason (Table~\ref{tab:decomp}), but our findings may not transfer to claims whose evidence we could not date. AVeriTeC and QuanTemp++ also use different dating instruments, so cross-dataset temporal comparisons are indicative, not controlled. On QuanTemp++ the outcome target $K$ is generated from fact-checker justification text rather than a neutral article summary (most AVeriTeC article failures use the same fallback), so the MI penalty on that corpus partly regularizes against verdict wording.

Third, our detector uses a single backbone (Qwen3-14B) and two datasets that differ a lot in how much post-claim evidence their retrieval surfaces, so the size of the temporal effect may not transfer to other corpora.

Fourth, the trained-detector results in Sections~\ref{sec:replevel} and~\ref{sec:temporal} operate on compressed features: the backbone states are mean-pooled and reduced with PCA, which keeps only part of the variance. There is a chance that our observation of the head not using the evidence may come from this compression rather than from the model.

\section{Future Work}
Two questions follow directly from these limitations. The first is whether a debiased detector holds up on claims made after the model's training cutoff, where memorization cannot help and an ordinary model should fall hard; no dataset with dated evidence exists for this yet, so it has to be collected. The second, and the more consequential one, is whether high-quality evidence exists before the claim at all, across many corpora: if misinformation usually spreads before good counter-evidence is published, every point-in-time system has a built-in ceiling, and measuring that ceiling matters more than any single detector.

\section{Conclusion}
\label{sec:conclusion}
Future-information leakage makes LLM-based misinformation detection look better than it really is, and it enters through two channels: memorized pre-training outcomes and retrieved documents written after the claim. We measure both on the same claims and find that width bottlenecks beat a mutual-information penalty on parametric leakage, while post-claim evidence measurably inflates a model that reads it on AVeriTeC. The two channels need different tools; neither alone is enough.

\bibliography{aaai2026}

\FloatBarrier
\clearpage
\noindent{\large\bfseries Appendix}\par\vspace{0.25\baselineskip}
\appendix
\setcounter{secnumdepth}{1}
\begingroup
\footnotesize
\section{Head configuration selection}
\label{app:configs}

For each dataset we sweep head configurations at $\lambda{=}0$, $d_R{=}32$,
with five seeds, and select the smallest train--validation gap that holds
validation F1. On AVeriTeC (Table~\ref{tab:cfg-avt}) linear heads generalize
best and E is selected. On QuanTemp++ (Table~\ref{tab:cfg-qt}) no
configuration overfits and D2 is selected.

\noindent\begin{minipage}{\columnwidth}
\centering
\footnotesize
\captionof{table}{Configuration sweep on AVeriTeC, chronological boundary. Configs
vary the feature PCA dimension $d_h$, the head hidden width (h), dropout
(drop), and weight decay (wd).}
\label{tab:cfg-avt}
\begin{tabularx}{\columnwidth}{@{}>{\raggedright\arraybackslash}Xcccc@{}}
\toprule
Config & train F1 & val F1 & gap & leakage \\
\midrule
D\phantom{2} $d_h{=}16$, h32, drop 0.3, wd $10^{-2}$ & 0.692 & 0.554 & $+0.137$ & 0.238 \\
D2 as D, wd $3{\times}10^{-2}$ & 0.683 & 0.551 & $+0.131$ & 0.232 \\
D3 as D, wd $10^{-1}$ & 0.668 & 0.532 & $+0.137$ & 0.222 \\
D4 as D, drop 0.5 & 0.674 & 0.555 & $+0.118$ & 0.238 \\
D5 as D4, wd $3{\times}10^{-2}$ & 0.673 & 0.545 & $+0.129$ & 0.234 \\
H\phantom{2} $d_h{=}8$, h16, drop 0.3, wd $10^{-2}$ & 0.626 & 0.522 & $+0.105$ & 0.164 \\
E\phantom{2} linear, $d_h{=}16$, wd $10^{-1}$ & 0.689 & 0.611 & $+0.078$ & 0.301 \\
F\phantom{2} linear, $d_h{=}16$, wd $10^{-2}$ & 0.672 & 0.594 & $+0.078$ & 0.301 \\
\bottomrule
\end{tabularx}
\end{minipage}

\noindent\begin{minipage}{\columnwidth}
\centering
\footnotesize
\captionof{table}{Configuration sweep on QuanTemp++, chronological boundary.}
\label{tab:cfg-qt}
\begin{tabularx}{\columnwidth}{@{}>{\raggedright\arraybackslash}Xcccc@{}}
\toprule
Config & train F1 & val F1 & gap & leakage \\
\midrule
D\phantom{2} $d_h{=}16$, h32, drop 0.3, wd $10^{-2}$ & 0.752 & 0.784 & $-0.032$ & 0.033 \\
D2 as D, wd $3{\times}10^{-2}$ & 0.749 & 0.789 & $-0.039$ & 0.032 \\
D3 as D, wd $10^{-1}$ & 0.742 & 0.776 & $-0.035$ & 0.031 \\
D4 as D, drop 0.5 & 0.747 & 0.787 & $-0.040$ & 0.034 \\
D5 as D4, wd $3{\times}10^{-2}$ & 0.745 & 0.777 & $-0.032$ & 0.032 \\
H\phantom{2} $d_h{=}8$, h16, drop 0.3, wd $10^{-2}$ & 0.720 & 0.751 & $-0.032$ & 0.024 \\
E\phantom{2} linear, $d_h{=}16$, wd $10^{-1}$ & 0.736 & 0.751 & $-0.015$ & 0.050 \\
F\phantom{2} linear, $d_h{=}16$, wd $10^{-2}$ & 0.727 & 0.759 & $-0.033$ & 0.050 \\
\bottomrule
\end{tabularx}
\end{minipage}

\section{Adjudication and evidence-amount details}
\label{app:details}

Cells flagged at 5 seeds are re-run with 20 paired seeds (Table~\ref{tab:adj}).
Figure~\ref{fig:knobs-claimonly} repeats the $\lambda$ sweep in the claim-only
setting where QuanTemp++ is the only pass. Table~\ref{tab:kcurve} supports the
AVeriTeC claim-only saturation at $d_R{=}8$, $\lambda{=}0$.

\begin{center}
\includegraphics[width=0.85\columnwidth]{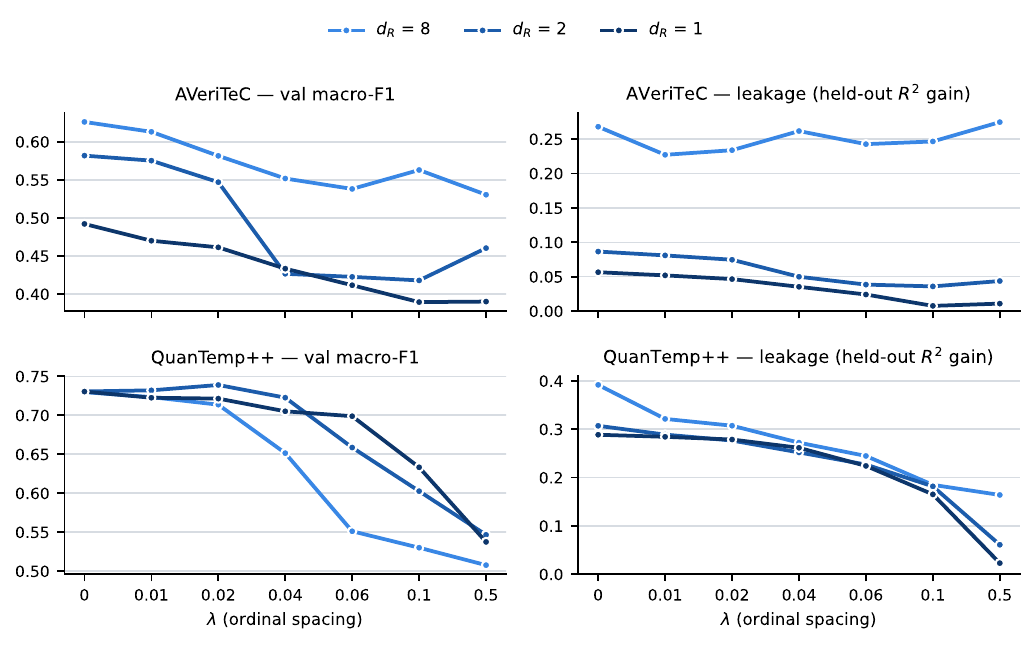}
\captionof{figure}{The $\lambda$ sweep of Figure~\ref{fig:rq2} in the claim-only setting.
With no evidence to condition on, the leakage term is unconditional.
QuanTemp++ is the one setting where a small $\lambda$ removes leakage at negligible F1
cost (Table~\ref{tab:adj}). Means over 5 seeds.}
\label{fig:knobs-claimonly}
\end{center}

\noindent\begin{minipage}{\columnwidth}
\centering
\footnotesize
\captionof{table}{20-seed paired adjudication of the candidate cells under each
dataset's selected configuration, all at $\lambda{=}0.01$ against the
$\lambda{=}0$ baseline.}
\label{tab:adj}
\begin{tabularx}{\columnwidth}{@{}>{\raggedright\arraybackslash}Xcc@{}}
\toprule
Arm & $\Delta$leakage [95\% CI] & $\Delta$F1 \\
\midrule
AVeriTeC top-5, $d_R{=}8$ & $-0.039$ $[-0.047, -0.031]$ & $-0.019$ \\
AVeriTeC top-5, $d_R{=}2$ & $-0.008$ $[-0.013, -0.004]$ & $-0.013$ \\
AVeriTeC claim-only, $d_R{=}2$ & $-0.010$ $[-0.015, -0.005]$ & $-0.011$ \\
QuanTemp++ top-5, $d_R{=}8$ & $-0.002$ $[-0.003, -0.001]$ & $-0.024$ \\
QuanTemp++ claim-only, $d_R{=}8$ & $-0.088$ $[-0.101, -0.075]$ & $-0.007$ \\
QuanTemp++ claim-only, $d_R{=}2$ & $-0.019$ $[-0.025, -0.014]$ & $-0.001$ \\
\bottomrule
\end{tabularx}
\end{minipage}

\noindent\begin{minipage}{\columnwidth}
\centering
\footnotesize
\captionof{table}{Evidence amount on AVeriTeC ($d_R{=}8$, $\lambda{=}0$): retrieval
saturates at one answer and claim-only is the highest-utility setting.}
\label{tab:kcurve}
\begin{tabular}{@{}lcc@{}}
\toprule
top-$k$ evidence & F1 & leakage \\
\midrule
0 (claim-only) & 0.626 & 0.268 \\
1 & 0.611 & 0.172 \\
2 & 0.609 & 0.169 \\
3 & 0.607 & 0.168 \\
5 & 0.607 & 0.168 \\
\bottomrule
\end{tabular}
\end{minipage}

\section{Evidence composition}
\label{app:composition}

\noindent\begin{minipage}{\columnwidth}
\centering
\captionof{table}{Leaked top-5 composition and overlap with the point-in-time view.}
\label{tab:composition}
\begin{tabular}{@{}lrr@{}}
\toprule
 & AVeriTeC & QuanTemp++ \\
\midrule
Binary claims                     & 2{,}513 & 5{,}537 \\
\quad post-claim, trusted date    & 7.7\%   & 5.8\%   \\
\quad post-claim, archive bound   & 40.1\%  & 0.0\%   \\
\quad pit-eligible (redundant)    & 31.9\%  & 90.8\%  \\
\quad undated                     & 20.2\%  & 3.4\%   \\
Claims with leaked $=$ pit top-5  & 19.4\%  & 70.4\%  \\
\bottomrule
\end{tabular}
\end{minipage}

\endgroup

\end{document}